\documentclass[lettersize,journal]{IEEEtran}
\usepackage{amsmath,amsfonts}
\usepackage{algorithmic}
\usepackage{algorithm}
\usepackage{array}
\usepackage[caption=false,font=normalsize,labelfont=sf,textfont=sf]{subfig}
\usepackage{textcomp}
\usepackage{stfloats}
 \usepackage{booktabs} 
\usepackage{url}
\usepackage{verbatim}
\usepackage{graphicx}
\usepackage{cite}
\usepackage{xcolor}
\definecolor{iccvblue}{rgb}{0.21,0.49,0.74}

\usepackage[pagebackref,breaklinks,colorlinks,allcolors=iccvblue]{hyperref}
\usepackage[capitalize]{cleveref}
  \makeatletter
  \newcommand{\onedot}{\@ifnextchar.{}{.}}
  \newcommand{\etal}{\emph{et al}\onedot}
\makeatother
\begin{document}

\title{Streamlining the Development of \\ Active Learning Methods in Real-World Object Detection}

\author{Moussa Kassem Sbeyti$^{1,2,3}$\quad Nadja Klein$^{1}$\\ Michelle Karg$^{3}$\quad Christian Wirth$^{3}$\quad Sahin Albayrak$^{2}$ \\
$^1$Scientific Computing Center, Karlsruhe Institute of Technology\\ $^2$DAI-Labor, TU Berlin \quad $^3$Continental AG \\  
\tt\small{\{moussa.sbeyti, nadja.klein\}@kit.edu}\quad\quad\tt\small{sahin.albayrak@dai-labor.de}\\
\tt\small{\{michelle.karg, christian.2.wirth\}@continental-corporation.com}\\ 

\thanks{This work has been funded by the KiKIT (The Pilot Program for Core-Informatics at the KIT) of the Helmholtz Association. It was also supported by the Helmholtz Association Initiative and Networking Fund on the HAICORE@KIT partition. Additionally, this work acknowledges financial support by the German Federal Ministry for Economic Affairs and Climate Action (BMWK) and the European Union within the project ``just better DATA (jbDATA)'', grant no. 19A23003A.}} 

\maketitle

\begin{abstract}
Active learning (AL) for real-world object detection faces computational and reliability challenges that limit practical deployment. Developing new AL methods requires training multiple detectors across iterations to compare against existing approaches. This creates high costs for autonomous driving datasets where the training of one detector requires up to 282 GPU hours. Additionally, AL method rankings vary substantially across validation sets, compromising reliability in safety-critical transportation systems.
We introduce object-based set similarity ($\mathrm{OSS}$), a metric that addresses these challenges. $\mathrm{OSS}$ (1) quantifies AL method effectiveness without requiring detector training by measuring similarity between training sets and target domains using object-level features. This enables the elimination of ineffective AL methods before training. Furthermore, $\mathrm{OSS}$ (2) enables the selection of representative validation sets for robust evaluation. 
We validate our similarity-based approach on three autonomous driving datasets (KITTI, BDD100K, CODA) using uncertainty-based AL methods as a case study with two detector architectures (EfficientDet, YOLOv3). 
This work is the first to unify AL training and evaluation strategies in object detection based on object similarity. $\mathrm{OSS}$ is detector-agnostic, requires only labeled object crops, and integrates with existing AL pipelines. This provides a practical framework for deploying AL in real-world applications where computational efficiency and evaluation reliability are critical. Code is available at \url{https://mos-ks.github.io/publications/}.
\end{abstract}

\section{Introduction}\label{sec:intro}
The effective deployment of object detectors in real-world applications relies on their ability to generalize across diverse and unfamiliar situations. While more data enhances generalizability, obtaining large, labeled, and diverse datasets is challenging due to redundancy, noise, and costly manual labeling. Active learning (AL) \cite{ren2021survey,garcia2023ten} aims to address this challenge by identifying samples that maximize learning while minimizing labeling effort. A typical pool-based AL pipeline consists of training a model on an initial labeled subset of the pool, predicting on the remaining unlabeled data, selecting the most informative samples for labeling, and retraining with the expanded labeled subset. Since no universal definition of informativeness exists, each AL method quantifies it differently. AL methods are evaluated based on detector performance trained on the selected samples. Therefore, informativeness in AL refers to the contribution of a sample to improving model performance when added to the training set.
\IEEEpubidadjcol

One common approach to quantifying informativeness is via uncertainty, which reflects the confidence of the model in its predictions \cite{roy2018deep,schmidt2020advanced,gao2020consistency,choi2021active}. In this work, we utilize widely used uncertainty-based AL methods as a case study for our $\mathrm{OSS}$ metric. Multiple works hypothesize the potential of separating between epistemic (model), aleatoric (data), localization, and classification uncertainties to increase the informativeness of a training set selected based on uncertainty \cite{kao2019localization,tagasovska2019single,venturini2020uncertainty,hullermeier2021aleatoric,kazemi2022complementing,nguyen2022measure,valdenegro2022deeper}. This is hereafter referred to as the \textit{separation hypothesis}. However, uncertainty correlates with informativeness only if accurately estimated, highlighting the importance of its \textit{calibration} \cite{pop2018deep,tang2021towards,hacohen2022active}. Furthermore, real-world datasets naturally exhibit class imbalance. Relying only on uncertainty for selection could amplify the imbalance and harm performance by prioritizing problematic samples regardless of their class. This underscores the significance of \textit{class balancing} \cite{brust2018active,kothawade2022talisman,wu2022entropy}. We therefore investigate the impact of uncertainty separation hypothesis, calibration, and class balancing on informativeness and their implications for real-world AL deployment.

Such an investigation comparing $m$ methods across $n$ iterations with $k$ training seeds would result in $k \times m \times n$ training runs. Training object detectors on real-world datasets usually requires many days on high-end GPUs, making AL method development for real-world applications computationally expensive. $\mathrm{OSS}$ quantifies the informativeness of a training set based on the data only. As a result, $\mathrm{OSS}$ can rank AL methods before any training occurs, enabling practitioners to eliminate ineffective AL methods early and focus computational resources on the most promising approaches.

Furthermore, uncertainty-based AL methods for object detection typically rely on two well-curated datasets \cite{haussmann2020scalable}: PASCAL VOC \cite{everingham2010pascal} and MS COCO \cite{lin2014microsoft}, with the evaluation conducted separately on each. However, when training extends beyond these benchmarks, inconsistencies emerge due to variations in training and evaluation strategies across previous works \cite{mittal2019parting,haussmann2020scalable,feng2022albench}. This leads to poor generalization to the more diverse and challenging real-world. Therefore, it is crucial to ensure the robustness of AL evaluation results \cite{haussmann2020scalable, kothawade2022talisman, feng2022albench, mittal2019parting}. 

\begin{figure}[ht]
  \centering
  \includegraphics[width=0.8\columnwidth]{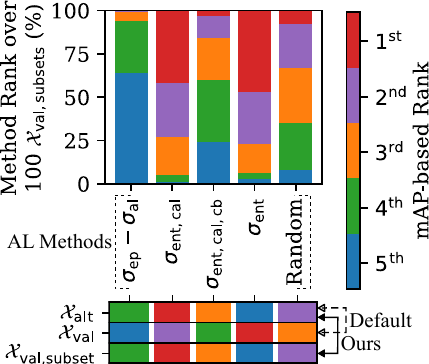} 
    \caption{
    The ranking of AL methods depends significantly on the composition of the validation dataset. This is illustrated for five AL methods (see \cref{sec:strat}) trained on 10\% of KITTI and evaluated on 100 different subsets $\mathcal{X}_{\mathrm{val,subsets}}$ of the KITTI validation set $\mathcal{X}_{\mathrm{val}}$ (top). Their ranking also varies when evaluated on CODA \cite{li2022coda}, an alternative evaluation set $\mathcal{X}_{\mathrm{alt}}$ with domain shift (bottom). In contrast, our approach identifies a representative $\mathcal{X}_{\mathrm{val,subset}}$ that maintains consistent ranking of AL methods across datasets, improving generalizability of evaluation results under domain shift.}
    \label{fig:introfig}
\end{figure}

Autonomous driving datasets exhibit significant domain variations due to geographic differences, weather conditions, traffic patterns, and infrastructure variations. An AL method that performs well on one validation subset may perform poorly on another, leading to unreliable deployment decisions. We visualize such inconsistencies in AL evaluation and motivate our approach in \cref{fig:introfig}. The evaluation results of five AL methods on a validation set $\mathcal{X}_{\text{val}}$ fail to generalize to its subsets or an alternative evaluation set $\mathcal{X}_{\text{alt}}$ with domain shift, as reflected in the ranking variations. For instance, if a dataset mostly contains daytime images with few nighttime samples, an unstratified validation set may contain only nighttime images. Even if an AL method yields a strong daytime detector, its validation performance will appear poor due to domain shift. We mitigate this by identifying a validation subset with $\mathrm{OSS}$ that better aligns with the target domain, providing a more reliable indicator of real-world performance than the original validation set (see \cref{fig:introfig}).

This paper introduces a novel similarity-based approach that addresses both the computational efficiency and evaluation reliability challenges in AL for real-world applications. We propose that the factors determining the informativeness of a training set also influence the representativeness of an evaluation set. We therefore propose unifying the AL training and evaluation strategies under object-based set similarity and introduce a novel metric to quantify it. 

Our key insight is that both AL method training set informativeness and evaluation set representativeness can be predicted by measuring how similar the sets are to the target domain. For this, $\mathrm{OSS}$ analyzes the distribution of visual features (aspect ratios, color histograms, and texture patterns) across object crops and their class distributions. As a result, $\mathrm{OSS}$ is model- and framework-agnostic, relying only on images and labels. 

In summary, our novel $\mathrm{OSS}$ streamlines the development of AL methods by avoiding the computational costs resulting from multiple training runs \cite{lowell2018practical,riedlinger2022towards,feng2022albench} while also improving the generalizability of the evaluation results, even to domain shift. $\mathrm{OSS}$ does not influence training sample selection. AL methods must still select samples based on their metric for informativeness, which motivates our case study using uncertainty-based AL methods on autonomous driving datasets. Therefore, our contributions cover the various stages of an AL pipeline, spanning from selection and training to evaluation and deployment:
\begin{itemize}
\item We establish that the informativeness of an AL training set correlates with its similarity to the data pool and the evaluation set, hence introducing an $\mathrm{OSS}$ metric that reliably predicts performance before training.
\item We evaluate uncertainty-based AL methods as a case study, including the impact of various types of uncertainties, class balancing, and calibration w.r.t.~similarity and informativeness on real-world datasets. 
\item We further utilize our $\mathrm{OSS}$ metric to improve the consistency of the AL evaluation results and their generalizability to the target domain despite domain shift.
\end{itemize}

\section{Related Work} \label{sec:relwork}
AL aims to identify the most informative samples in the data pool, although recent works diverge on the definition of informativeness. Some assert that it translates to selecting samples with the highest predictive uncertainty under the assumption that samples that provide new information to the model will highlight its limitations \cite{roy2018deep,schmidt2020advanced,choi2021active, haussmann2020scalable}. Others consider diversity since diverse samples generally contain a broader spectrum of information \cite{hacohen2022active,tharwat2023survey}. Alternative definitions include difficulty via loss prediction on unlabeled data \cite{yoo2019learning}, localization disagreement between anchor boxes \cite{roy2018deep}, and localization stability post-corruption \cite{kao2019localization}. 

\textbf{Uncertainty vs.~Informativeness:} When assessing uncertainty-based AL, several works propose separating localization and classification uncertainties \cite{choi2021active,yoo2019learning,kao2019localization}. Hacohen \etal~\cite{hacohen2022active} observe that uncertain classification examples offer valuable insights, provided the uncertainty is accurately estimated, highlighting the importance of calibration \cite{pop2018deep,tang2021towards}. To address high uncertainty introduced by commonly challenging features such as crowded or nighttime scenes, Tang \etal~\cite{tang2021towards} incorporate diversity through clustering based on feature maps. However, Kothawade \etal~\cite{kothawade2022talisman} argue that diversity and uncertainty may be ineffective at selecting rare classes. This is especially relevant in autonomous driving datasets with imbalanced classes such as cars vs.~motorcycles. Class balancing can mitigate this through class-based weighting of the loss function \cite{yu2022re,chawla2002smote,schmidt2020advanced}, class-based weighting of the AL selection function with the number of instances and classes \cite{brust2018active,wu2022entropy}, and considering the distance between class distributions in the selected and labeled sets \cite{yu2022consistency}. Informativeness is therefore influenced by several factors, which we further investigate on real-world datasets.

\textbf{Quantifying Informativeness:} Broadening the scope, Yu \etal~\cite{yu2022consistency} argue that a valid metric for informativeness ensures that information decreases over AL iterations. Furthermore, the information in the selected set of each iteration should surpass that in the labeled set to enhance performance. However, they do not propose an approach to quantify the information content within a set. We address this by linking informativeness to similarity and introducing a novel $\mathrm{OSS}$ metric. One relevant work on similarity addresses the selection of instances specific to rare events \cite{kothawade2022talisman}. Kothawade \etal~\cite{kothawade2022talisman} use submodular mutual information functions, which measure the amount of information in the data, to find images that are semantically similar to regions of interest in the query set. Yuan \etal~\cite{yuan2021multiple} address distribution bias between labeled and unlabeled instances, thereby increasing their similarity, by minimizing prediction discrepancies and aligning features through an adversarial classifier setting. These approaches target sample informativeness and sample selection. In contrast, our approach measures set similarity post-selection based solely on images and corresponding labels. It therefore supports the development of AL selection methods and is efficient and well-suited for real-world object detection. 

\textbf{Evaluation Reliability:} Previous works on AL have primarily focused on performance metrics rather than evaluation reliability. Mittal \etal~\cite{mittal2019parting} advocate for a more robust evaluation strategy in AL that encompasses a broader spectrum of datasets, ranging from low- to large-budget regimes. We reinforce and uphold this viewpoint by validating our findings on datasets differing in size by a factor of 10. Nevertheless, acquiring diverse datasets poses a challenge. Therefore, we utilize $\mathrm{OSS}$ to address inconsistent evaluation \cite{haussmann2020scalable, mittal2019parting} and enhance reliability without requiring additional data collection or labeling of more evaluation sets. $\mathrm{OSS}$ helps identify a subset of the evaluation set that is more representative of the target domain, which is crucial for the dynamic environment of real-world applications such as autonomous driving.

\section{Leveraging Similarity in Active Learning} \label{sec:simal}
This section establishes the theoretical foundation for our approach in three parts: we first formalize the AL setup (\cref{{subsec:AL}}), then demonstrate how similarity is linked to training set informativeness (\cref{sec:simalinfo}) and evaluation set representativeness (\cref{sec:simalrep}) in AL, and finally introduce our $\mathrm{OSS}$ metric that quantifies these relationships (\cref{sec:mtrc}).
\subsection{Active Learning Setup} \label{subsec:AL}
Pool-based AL follows an iterative setting on a pre-defined labeling budget $b\in\mathbb{N}_{>0}$ with a number of iterations $n_i \in\mathbb{N}_{>0}$,  $n_i\leq b$. It aims to find a subset of size $b$ that contains the most informative samples of a pool $\mathcal{D_{\text{pool}}}$. Given $\mathcal{D_{\text{pool}}}$ of size $n_{\text{pool}}\geq b$, we denote the labeled subset at iteration $i$ for $i=1,\ldots,n_i$, as $\mathcal{X}_l^i \subseteq \mathcal{D_{\text{pool}}}$ with the corresponding labels $\mathcal{Y}_l^i$ and the unlabeled subset at iteration $i$ as $\mathcal{X}_u^i \subseteq \mathcal{D_{\text{pool}}}$ such that $\mathcal{D_{\text{pool}}} = \mathcal{X}_l^i \bigcup \mathcal{X}_u^i$. Formally, each AL iteration $i$ consists of predicting on $\mathcal{X}_u^i$, selecting the $\lfloor \frac{b}{n_i} \rfloor\in\mathbb{N}_{>0}$ most informative samples for the model from $\mathcal{X}_u^i$, labeling them by an oracle, and adding them to $(\mathcal{X}_l^i, \mathcal{Y}_l^i)$, which results in $(\mathcal{X}_l^{i+1}, \mathcal{Y}_l^{i+1})$ as the new training set at iteration $i+1$. The validation set denoted by $(\mathcal{X}_{\text{val}}, \mathcal{Y}_{\text{val}})$ is selected and labeled from $\mathcal{D_{\text{pool}}}$ at the start and remains fixed.

\subsection{Similarity Link} \label{sec:subset}
We observe that the informativeness of an AL training set and the representativeness of an evaluation set are linked to similarity, as detailed in \cref{sec:subset}. Therefore, we introduce a novel object-based set similarity ($\mathrm{OSS}$) metric in \cref{sec:mtrc}. 

\subsubsection{Linking Similarity to Informativeness} \label{sec:simalinfo}
We propose that the objective in AL of identifying the most informative subset can be framed as finding the subset that maximizes similarity to $\mathcal{D_{\text{pool}}}$: $\max(\mathrm{OSS}(\mathcal{X}_l\|\mathcal{D_{\text{pool}}}))$. Object detectors are trained to localize and classify specific objects. We contend that similarity quantification should be confined to these objects, therefore requiring labeled data. 

Since the majority of $\mathcal{D_{\text{pool}}}$ is unlabeled, $\mathrm{OSS}(\mathcal{X}_l\|\mathcal{D_{\text{pool}}})$ is approximated by $\mathrm{OSS}(\mathcal{X}_l\|\mathcal{X}_{\text{val}})$. The performance of an AL method is ultimately measured on $\mathcal{X}_{\text{val}}$, typically via the mean Average Precision (mAP) for object detection \cite{lin2014microsoft}. Therefore, the mAP at an iteration $i$ should positively and linearly correlate with $\mathrm{OSS}(\mathcal{X}_l^i\|\mathcal{X}_{\text{val}})$, emphasizing the need for an $\mathcal{X}_{\text{val}}$ that is representative of $\mathcal{D_{\text{pool}}}$. 

\subsubsection{Linking Similarity to Representativeness} \label{sec:simalrep}
A well-defined similarity metric should not only predict training set informativeness before training, but also support the identification of an evaluation set representative of the target domain. For example, if the class ``Pedestrian'' is infrequent in $\mathcal{X}_{\text{val}}$ but is prevalent in $\mathcal{D}_{\text{pool}}$, or if its characteristics differ significantly between sets, the mAP-based ranking of AL methods on $\mathcal{X}_{\text{val}}$ may not reflect their general effectiveness. 

Since $\mathrm{OSS}(\mathcal{X}_l^i\|\mathcal{X}_{\text{val}})$ correlates with performance, we assert that the similarity between $\mathcal{X}_{\text{val}}$ and an alternative evaluation set $\mathcal{X}_{\text{alt}}$ reflects the reliability of the performance-based ranking of AL methods on $\mathcal{X}_{\text{val}}$. Here, $\mathcal{X}_{\text{alt}}$ can be a subset of $\mathcal{X}_{\text{val}}$ itself, an $\mathcal{X}_{\text{val}}$ from another dataset, or a test set $\mathcal{X}_{\text{test}}$ depending on the AL setup. Thus, the representativeness of $\mathcal{X}_{\text{val}}$ is linked to its similarity to other evaluation sets, suggesting greater generalizability to $\mathcal{D}_{\text{pool}}$. 

\textbf{Evaluation Set Identification:} We propose that a more reliable evaluation is achieved by identifying a subset $\mathcal{X}_{\text{val,subset}} \subseteq \mathcal{X}_{\text{val}}$ with $\mathrm{OSS}(\mathcal{X}_{\text{val}}\|\mathcal{X}_{\text{alt}}) \leq \mathrm{OSS}(\mathcal{X}_{\text{val,subset}}\|\mathcal{X}_{\text{alt}})$ via subsampling before starting the AL iterations. Let $\mathcal{X}_{\text{val,subset}}^1,\ldots, \mathcal{X}_{\text{val,subset}}^z$ be randomly drawn, equally-sized subsets of $\mathcal{X}_{\text{val}}$, with $z$ manually determined based on the size of $\mathcal{X}_{\text{val}}$. The subset with the highest similarity to $\mathcal{X}_{\text{alt}}$, given by $\mathrm{OSS}_z = \max_{l \in {1, \ldots, z}} \mathrm{OSS}(\mathcal{X}_{\text{val,subset}}^l \| \mathcal{X}_{\text{alt}})$, is selected as the new validation set. This approach requires no additional labeling since representative subsets can be identified from existing evaluation sets. The validation sets remain fixed across AL iterations, so their superscripts differ from those of the training sets. 

\begin{figure*}[ht]
  \centering
  \includegraphics[width=\textwidth]{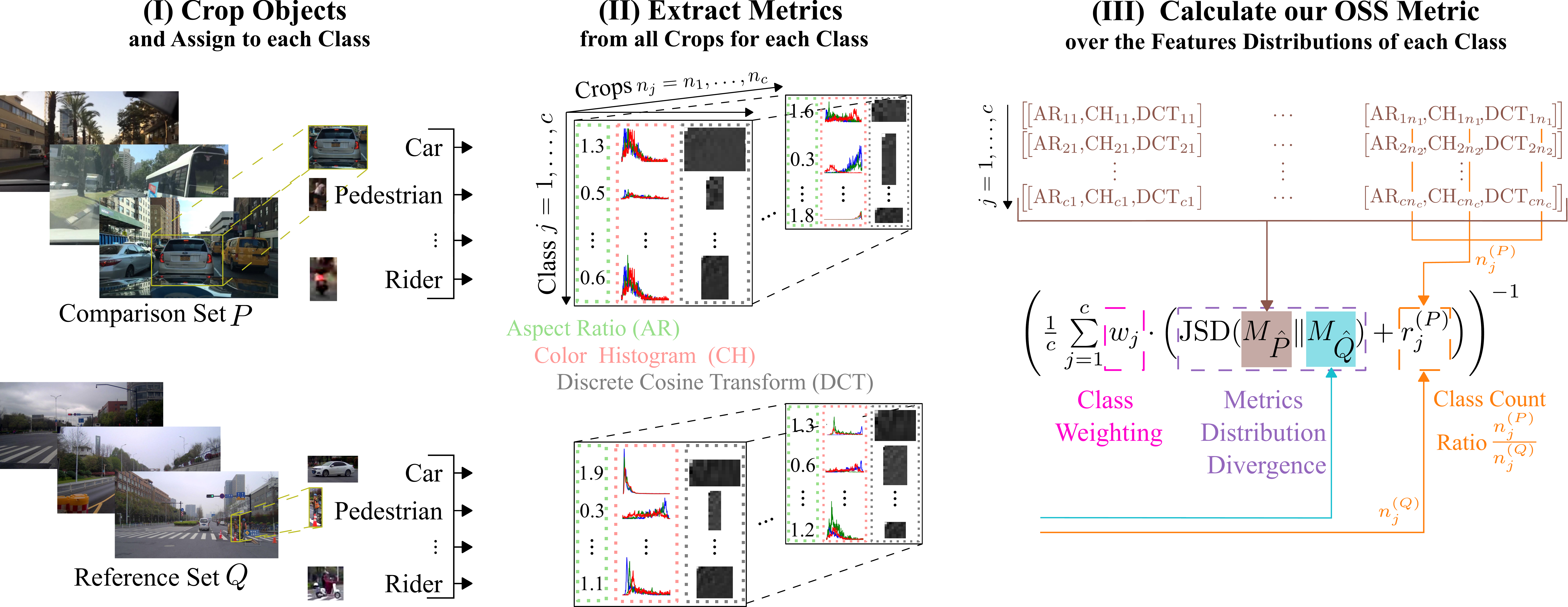}
    \caption{$\mathrm{OSS}$ quantifies both the informativeness of a training set and the representativeness of an evaluation set. By measuring the similarity between selected training sets ($P$) and a validation set ($Q$), it assesses the effectiveness of AL methods without training. Hence, it reduces computational costs by eliminating ineffective AL methods. Additionally, by measuring the similarity between validation subsets ($P$) and an alternative set ($Q$), it helps identify a representative evaluation set that mitigates biases in evaluation results and improves generalization to $\mathcal{D}_{\text{pool}}$. To calculate $\mathrm{OSS}$ as proposed in \cref{eq:goeq}, (\textbf{I}) object crops for each class are first extracted. (\textbf{II}) Three features are then calculated as descriptors of object characteristics across spatial and frequency domains. (\textbf{III}) The inverse of the divergence between the features distributions combined with the class count ratio for each class within the two sets serves as a similarity measure between them. }
    \label{fig:gofig}
\end{figure*} 

\subsection{Object-based Set Similarity ($\mathrm{OSS}$) Metric}\label{sec:mtrc}
Having established the theoretical links between similarity, informativeness, and representativeness, we now introduce the $\mathrm{OSS}$ metric that translates these theoretical insights into a practical similarity measure.
\subsubsection{Design Principles} 
We introduce a novel similarity metric based on the two links in \cref{sec:subset}. The process is visualized in \cref{fig:gofig}.
\begin{itemize}
    \item \textbf{Object-Centric Analysis:} Real-world datasets often contain redundant information within and across images. Detectors are typically evaluated using mAP, which averages detection performance per class. We therefore measure similarity based on the object crops per class $j=1,\ldots,c$, prioritizing relevant features while reducing computation time due to lower resolution. We calculate similarity between two sets of images $P$ (the comparison set) and $Q$ (the reference set) by (\textbf{I}) cropping the images within these sets. This enables object-centric analysis focused on detection-relevant content.
    \item \textbf{Multi-Modal Feature Representation:} Real-world objects exhibit diverse visual characteristics requiring combined spatial, frequency, and color analysis. For each class $j$, we (\textbf{II}) extract three complementary features from object crops: 
    \begin{itemize}
        \item Aspect ratio (AR) captures object shape characteristics critical for real-world applications, e.g., vehicle orientations and pedestrian poses vary significantly.
        \item Mean of the 2D discrete cosine transform (DCT) \cite{ahmed2006discrete} coefficients quantifies texture patterns that distinguish object classes.
        \item Mean value of the flattened 3D color histogram (CH) represents color distributions affected by lighting conditions, weather, and object materials.
    \end{itemize}
    In summary, objects of each class are encoded via complementary features covering color, texture, and shape.
    \item \textbf{Class-Balanced Evaluation:} Real-world data exhibits class imbalance requiring weighted similarity computation. We (\textbf{III}) calculate the class-weighted multivariate Jensen-Shannon divergence (JSD) \cite{duchi2007derivations} over the empirical distributions of the features and the class count ratio between the two sets to achieve class-balanced evaluation.
\end{itemize}

\subsubsection{Similarity Computation}
The $\mathrm{OSS}$ metric combines distributional divergence, class count ratios, and class-based weighting. 

\textbf{Distribution Divergence:} JSD requires equal sample sizes from $P$ and $Q$, which may not hold (e.g., different sizes of training and evaluation sets). We therefore use Gaussian Kernel Density Estimation (KDE) \cite{scott2015multivariate,scipy} to model their probability density functions \( \hat{f}_P(x) \) for \( P \) and \( \hat{f}_Q(x) \) for \( Q \). We then draw equal-sized samples, forming \( \hat{P} \) and \( \hat{Q} \). Let \( M_{\hat{P}} \in \mathbb{R}^{c \times 3} \) and \( M_{\hat{Q}} \in \mathbb{R}^{c \times 3} \) denote the feature matrices calculated for \( \hat{P} \) and \( \hat{Q} \), respectively. Each row corresponds to an object class, and each column to one of the three features. The JSD is estimated directly on $M_{\hat{P}}$ and $M_{\hat{Q}}$ as outlined in Eq.~14 of \cite{perez2008kullback}, a practical non-parametric estimation between continuous distributions without assuming a restrictive parametric form. 

\textbf{$\mathrm{OSS}$ Formulation:} To compare multiple AL methods simultaneously, we calculate $\mathrm{OSS}(P_m \| Q)\in \mathbb{R}^+$ in \cref{eq:goeq} between multiple comparison sets $P_m$ and $Q$, $m=1,\ldots,k$, and $k\geq1$ as:

\begin{multline} \label{eq:goeq} 
     \mathrm{OSS}(P_m \| Q) =\\ \left( \frac{1}{c}\sum\limits^{c}_{j=1} w_j \cdot \left(\text{JSD}(M_{\hat{P}_m} \| M_{\hat{Q}}) + r_j^{(P_m)} \right) \right)^{-1}.
\end{multline}

\textbf{Class Count:} The class count ratio $r_j^{(P_m)}$ accounts for training set scalability, as more object samples typically yield a more representative set. It is defined as the smoothed ratio of the class counts $n_j^{(P_m)}$ and $n_j^{(Q)}$ for a class $j$, that is,
\begin{equation}\label{eq:reg}
\begin{aligned}  
    r_j^{(P_m)} &=  \frac{1}{4}\cdot\beta^{(P_m)} \cdot\tfrac{n_j^{(P_m)}}{n_j^{(Q)}} + \frac{1}{2}.
\end{aligned} 
\end{equation}
For smoothing, we use the first-degree Taylor polynomial $T_1(x) = \frac{1}{2} + \frac{1}{4}x$ as a linear approximation of the sigmoid function at $x=0$. This approximation retains sensitivity to ratio changes while ensuring numerical stability and preventing extreme values \cite{bishop1995neural}. 

\textbf{Scale Normalization:} To address significant detection count variations among the $k$ comparison sets, $\beta^{(P_m)}$
normalizes $n_\text{det}^{(P_m)}$ by the 25th percentile ($Q_1$) of $n_\text{det}^{(P_m)}$ across all $k$ sets, denoted $Q_{1_{m=1}^{k}}(n_{\text{det}}^{(P_m)})$ in:
\begin{equation*}
\begin{aligned}  
    \beta^{(P_m)} &= \max\left(1, \frac{n_\text{det}^{(P_m)}}{Q_{1_{m=1}^{k}}(n_\text{det}^{(P_m)})}\right).
\end{aligned} 
\end{equation*}

\textbf{Imbalance Compensation:} Since class imbalance affects the similarity metric, we introduce a weighting factor to ensure each class contributes proportionally to its frequency in scenarios with substantial class imbalances. The class weight $w_j$ represents the average class proportion across the $k$ comparison sets:
\begin{equation} \label{eq:w}
w_{j} =
\begin{cases}
    1 & \text{if } \text{CV}_\mathbf{w} \leq C_w \\
       \frac{1}{k}\sum\limits_{m=1}^{k} \frac{n_j^{(P_m)}}{n_\text{det}^{(P_m)}} & \text{otherwise.}
\end{cases}
\end{equation} 
$\text{CV}_\mathbf{w}=\frac{\sigma_\mathbf{w}}{\mu_\mathbf{w}}$ is the coefficient of variation, with $\textbf{w} = (w_1, w_2, \ldots, w_c)^\top$, the mean $\mu_\mathbf{w}$, and the standard deviation $\sigma_\mathbf{w}$, and the manually defined activation threshold $C_w$. In case of activation, classes with a small weight $w_{j} < Q_{1_{j=1}^{c}}(w_j)$ are filtered out due to their negligible and unreliable contribution to the similarity metric.

\section{Experiments}
\subsection{Implementation Details} \label{sec:strat}
\subsubsection{Experimental Setup} We select EfficientDet-D0 \cite{tan2020efficientdet,automl} pre-trained on COCO \cite{lin2014microsoft} as the main baseline detector and three autonomous driving datasets for our experiments. These are KITTI (7 classes, 20\% random split for validation) \cite{geiger2012we} and BDD100K (10 classes, 12.5\% official split) \cite{yu2020bdd100k} for fine-tuning ($\mathcal{X}_l$ and $\mathcal{X}_u$) and evaluation ($\mathcal{X}_{\text{val}}$). The third dataset, CODA (4 classes in common with KITTI and 8 with BDD) \cite{li2022coda}, is our alternative evaluation set ($\mathcal{X}_{\text{alt}}$) with domain shift. Each AL iteration consists of a training from scratch for 200 epochs with 8 batches and an input resolution of 1024$\times$512 pixels. All other hyperparameters of EfficientDet maintain their default value \cite{tan2020efficientdet}. We initiate all iterations on a random 5\% of $\mathcal{D_{\text{pool}}}$ as a warm-up \cite{settles2009active}, corresponding to 299 images for KITTI (reasonable minimum covering all classes) and 3,492 images for BDD. Results represent four training iterations (two identical and two distinct warm-up seeds) as minimal variations are observed. 

\subsubsection{Uncertainty Estimation} For the case study, we estimate epistemic classification ($\sigma_\text{ep,cls}$) and localization ($\sigma_\text{ep,loc}$) uncertainties via 2D spatial Monte Carlo (MC) dropout \cite{tompson2015efficient} with a dropout rate of 0.05 selected based on best performance and 10 MC samples, cf. \cite{stoycheva2021uncertainty}. Higher dropout rates significantly reduce performance, e.g., increasing the rate to 0.1 reduces mAP by $3.49 \pm 0.38$. Our dropout rate and number of samples were also selected based on the low expected calibration error (ECE) of $0.11 \pm 0.02$ for $\sigma_\text{ep,loc}$. To estimate aleatoric uncertainty ($\sigma_\text{al}$), we use loss attenuation (LA) \cite{kendall2017uncertainties, sbeyti2023overcoming} in the localization head \cite{sbeyti2023overcoming}. We extract and apply softmax on the predicted classification logits to calculate entropy ($\sigma_\text{ent}$) as $\sigma_{\text{ent}}= -\sum_{j=1}^cp_j \log_2(p_j)$ over the confidences $p_j$ of the $c$ classes. We use isotonic regression per-class to calibrate $\sigma_\text{cls}$ and per-class and per-coordinate for $\sigma_\text{loc}$ \cite{sbeyti2023overcoming}, and denote them with the subscript $_\text{cal}$. $\sigma_\text{loc}$ is also normalized by the width or height depending on whether it pertains to an $x$- or $y$-coordinate. The uncertainties per object are averaged as $\sigma_{\text{loc}} = \frac{1}{4}\sum_{i=1}^4\sigma_{\text{loc},i}$ and $\sigma_{\text{cls}} = \frac{1}{c}\sum_{i=1}^c(\sigma_{\text{cls}, i})$.

\subsubsection{AL methods} We consider as baselines entropy-based sampling \cite{roy2018deep}, random sampling, and the widely adopted probabilistic approach \cite{choi2021active}. The latter standardizes all uncertainties and selects the maximum value per image. We compare these baseline AL methods to methods that cover the intuitions described in \cref{sec:intro,sec:relwork}. These include $\sigma_\text{ep,cls}$, $\sigma_\text{ep,loc}$, $\sigma_{\text{al}}$, $\sigma_\text{ep}=\sigma_\text{ep,cls}+\sigma_\text{ep,loc}$, $\sigma_\text{ep}- \sigma_{\text{al}}$, $\sigma_\text{ent}+\sigma_\text{al}$, and a calibrated version for each uncertainty $\sigma$. We apply min-max scaling to each $\sigma$ across all images so that all types of $\sigma$ $\in[0,1]$. We \textit{aggregate} the uncertainties per image using maximum across all objects, as we find that it outperforms both the mean and sum. We then apply top-$k$ selection across all images to select the most informative images for the next AL iteration \cite{haussmann2020scalable}. Additionally, we test the $n$-bins formulation with $n=5$ by dividing the score space into equal-sized bins and then querying from the top ($n$-1) bins (exploration) and the last bin (exploitation) \cite{roy2018deep}. We address class imbalance by weighting uncertainties (subscript $_\text{cb}$) according to the inverse frequency of each corresponding class in the predictions on $\mathcal{X}_u^i$, thereby prioritizing underrepresented classes \cite{brust2018active}.

\subsubsection{Evaluation Metrics}
\begin{itemize}
    \item \textbf{Correlation Analysis:} Pearson correlation coefficient $r$ \cite{pearson1895vii} and the p-value\footnote{Asterisks indicate the level of statistical significance: no asterisk $(\text{p} > 0.05)$, $^* (\text{p} \leq 0.05)$, $^{**} (\text{p} \leq 0.01)$, $^{***} (\text{p} \leq 0.001)$.}\label{sigfootnote} measure the linear relationship between $\mathrm{OSS}$ and mAP performance across AL methods.
    \item \textbf{Ranking Consistency:} Kendall's tau ($\tau$) \cite{kendall1938new} measures the consistency of mAP-based AL method rankings across multiple evaluation sets.
\end{itemize}

The following subsections present our experimental results in four parts: we first validate the similarity-informativeness and similarity-representativeness relationships in AL, followed by a comprehensive case study using uncertainty-based AL methods, and finally provide detailed ablation studies of the $\mathrm{OSS}$ components and its detector agnosticism.

\subsection{Linking Similarity to Informativeness}
\begin{figure}[ht]
  \centering
  \includegraphics[width=\columnwidth]{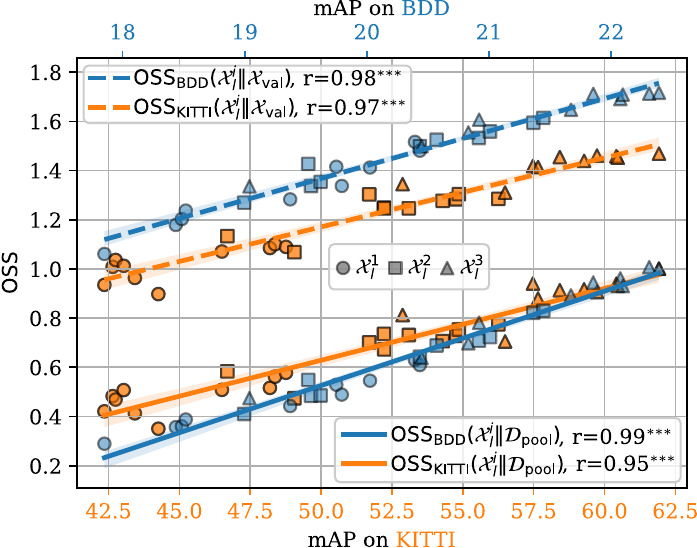}
    \caption{Strong linear correlation visualized and quantified via $r$ between performance (mAP) and $\mathrm{OSS}$, calculated for 10 AL methods across three AL iterations ($\mathcal{X}_l^1$ at 10\%, $\mathcal{X}_l^2$ at 15\%, and $\mathcal{X}_l^3$ at 25\% of $\mathcal{D}_{\mathrm{pool}}$). Our metric $\mathrm{OSS}(\mathcal{X}_l^i \| \mathcal{X}_{\mathrm{val}})$ consistently serves as an effective estimator of informativeness.}

    \label{fig:simap}
\end{figure}

We first demonstrate that similarity is linked to informativeness (see \cref{sec:simalinfo}). For 10 AL methods from \cref{sec:strat} at iterations $i=1,2,3$ in \cref{fig:simap}, a strong linear correlation exists between the mAP on $\mathcal{X}_{\mathrm{val}}$ and $\mathrm{OSS}(\mathcal{X}_l^{1-3} \| \mathcal{D_{\text{pool}}})$ and $\mathrm{OSS}(\mathcal{X}_l^{1-3} \| \mathcal{X}_{\mathrm{val}})$. This holds for both the randomly selected $\mathcal{X}_{\mathrm{val}}$ of KITTI and the curated one of BDD. The correlation is particularly strong for larger subsets (smallest deviation at $i=3$) but also exists despite a modest subset size of 598 images from KITTI (10\%) at $i=1$ and low disparity in mAP (below 1\%).  

The same experiments with CODA ($\mathrm{OSS}(\mathcal{X}_l^i \| \mathcal{X}_{\mathrm{alt}})$) yield an average
$r=0.85^{***}$\hyperref[sigfootnote]{\footnotemark[\value{footnote}]} for the three iterations on BDD and a weaker $r=0.27$\hyperref[sigfootnote]{\footnotemark[\value{footnote}]} on KITTI due to stronger class distribution shifts and fewer classes between KITTI and CODA. Additionally, CODA includes a larger proportion of BDD and contains night images, unlike KITTI \cite{li2022coda}.

\cref{tab:simvsap} extends the analysis towards per-class AP, showing a strong linear correlation with the AP despite an even lower deviation in AP ($<0.2$\%) and count (0.4\% of $\mathcal{X}_{\mathrm{val}}$). However, correlation strength varies based on class distinctiveness, with less descriptive classes (e.g., ``Traffic Sign'') exhibiting weaker correlation across comparison sets.

\begin{table}[ht]
\centering
\caption{KITTI (top) and BDD (bottom): Correlation coefficient $r$ between $\mathrm{OSS}$ and the AP for 10 AL methods at the first AL iteration (10\%), including the corresponding class proportion (\%).}
\label{tab:simvsap}
\resizebox{\columnwidth}{!}{%
\begin{tabular}{lcccccccccc}
&\rotatebox{75}{Car} & \rotatebox{75}{Van} & \rotatebox{75}{Truck} & \rotatebox{75}{\shortstack{Pedest- \\ rian}}& \rotatebox{75}{\shortstack{Person \\ Sitting}} & \rotatebox{75}{Cyclist} & \rotatebox{75}{Tram} &  &  &  \\
\midrule
 \% & 62,4 & 7.7 & 3.2 & 17.2 & 1.5 & 5.3 & 2.8 &  &  &  \\
$r\uparrow$& $0.90^{***}$ & $0.70^*$ & $0.87^{***}$ & $0.91^{***}$ & $0.89^{***}$ & $0.70^*$ & $0.96^{***}$ &  &  &  \\
 \midrule
&\rotatebox{75}{Car} & \rotatebox{75}{Rider} & \rotatebox{75}{Truck} & \rotatebox{75}{\shortstack{Pedest- \\ rian}} & \rotatebox{75}{\shortstack{Motor- \\ cycle}} & \rotatebox{75}{Bicycle} & \rotatebox{75}{Bus} & \rotatebox{75}{\shortstack{Traffic \\ Light}} & \rotatebox{75}{\shortstack{Traffic \\ Sign}}\\
\midrule
\% & 51.3 & 0.6 & 2.4 & 9.0 & 0.4 & 0.9& 1.1 & 15.6 & 18.7\\
$r\uparrow$& $0.87^{***}$ & $0.97^{***}$ & $0.88^{***}$ & $0.95^{***}$ & $0.97^{***}$ & $0.97^{***}$ & $0.85^{**}$ &0.47&0.10\\
 \bottomrule
\end{tabular}%
}
\end{table}

\textbf{Computational Cost Analysis:} In summary, $\mathrm{OSS}(\mathcal{X}_l^i \| \mathcal{X}_{\mathrm{val}})$ serves as an estimator of informativeness of an AL training set $\mathcal{X}_l^i$, since informativeness in AL refers to how much a set increases model performance (see \cref{sec:intro}). $\mathrm{OSS}$ requires no additional labeling and is computationally efficient, with a processing time of 1 minute for $\mathrm{OSS}(\mathcal{X}_l^1 \| \mathcal{X}_{\mathrm{val}})$ on KITTI and 12 minutes for 100 times more crops in BDD on an Intel Xeon Platinum 8168 CPU. In contrast, the training for a single AL method on BDD (six iterations) requires up to around 34 days on an NVIDIA Tesla V100 GPU (806 GPU hours), highlighting the time-saving advantage of our approach. $\mathrm{OSS}$ can be used to eliminate ineffective AL methods at each iteration before training, ensuring that only the most promising ones proceed to subsequent iterations. We demonstrate the cost benefits of the latter in our case study.

\subsection{Linking Similarity to Representativeness} We further demonstrate that similarity positively correlates with representativeness (see \cref{sec:simalrep}). An evaluation set is considered representative if the performance-based ranking of AL methods on it generalizes across other evaluation sets and ultimately to $\mathcal{D_{\text{pool}}}$. 

We first split the evaluation sets into 100 subsets with $1/10$ of the original size each. Next, we compute $\tau$ between the ranking of the five AL methods in \cref{fig:introfig} based on $\mathrm{OSS}$ and those based on the mAP. We select $\tau$ as we transition from measuring the linear correlation between the continuous values $\mathrm{OSS}$ and mAP to comparing the rankings of the AL methods. A $\tau$ value of 1.0 indicates complete agreement in ranks, while a $\tau$ value of -1.0 indicates complete disagreement. We analyze 100 subsets as well as the original $\mathcal{X}_{\mathrm{val}}$ and $\mathcal{X}_{\mathrm{alt}}$ (CODA). 

\begin{figure}[ht!]
  \centering
  \includegraphics[width=\columnwidth]{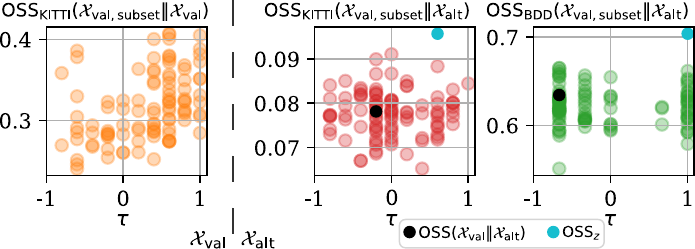}\caption{OSS vs.~$\tau$. $\tau$ measures the consistency of mAP-based ranking of AL methods across two evaluation sets. The ranking of AL methods on the 100 subsets of $\mathcal{X}_{\mathrm{val}}$ (colored circles) vary significantly ($\tau \in [-1.0,1.0]$ when rankings are compared to $\mathcal{X}_{\mathrm{val}}$ and $\mathcal{X}_{\mathrm{alt}}$). Moreover, the rankings on $\mathcal{X}_{\mathrm{val}}$ vs.~$\mathcal{X}_{\mathrm{alt}}$ exhibit a negative correlation, rendering the evaluation on $\mathcal{X}_{\mathrm{val}}$ unreliable. However, our method identifies a representative subset ($\mathrm{OSS}_z$) of $\mathcal{X}_{\mathrm{val}}$ with a consistent ranking across evaluation sets despite domain shift ($\tau = 0.6$ on KITTI and $\tau = 1.0$ on BDD).}
    \label{fig:evalall}
\end{figure}

\textbf{Evaluation Reliability Analysis:} First, $\mathrm{OSS}$ correlates with $\tau$ (\cref{fig:evalall} left), even despite the domain shift in $\mathcal{X}_{\mathrm{alt}}$ (right) and the sensitivity of $\tau$ to similar performance ($<0.2\%$ mAP difference). Second, the performance of the AL methods on one evaluation set is unreliable (strong fluctuation of $\tau$ in \cref{fig:evalall}). Third, as depicted in \cref{fig:introfig}, the majority rank on the 100 $\mathcal{X}_{\mathrm{val,subsets}}$ may not reflect the performance on $\mathcal{X}_{\mathrm{val}}$ and $\mathcal{X}_{\mathrm{alt}}$. 

 \textbf{Subsampling Analysis:} However, a subset may exist within $\mathcal{X}_{\mathrm{val}}$ where the ranking exactly matches that of $\mathcal{X}_{\mathrm{val}}$ and even $\mathcal{X}_{\mathrm{alt}}$ (visualized in \cref{fig:introfig} and \cref{fig:evalall}). This representative subset is directly identified based on its similarity ($\mathrm{OSS}_z$) to the reference set $\mathcal{X}_{\mathrm{val}}$ or $\mathcal{X}_{\mathrm{alt}}$, hence reducing the computational expense of evaluating on multiple subsets for a robust evaluation. We observe that subsampling a subset with higher similarity across sets improves the consistency of evaluation results when sampling and labeling more data is costly. Our similarity-based approach increases $\tau_\text{KITTI}(\mathcal{X}_{\mathrm{val}},\mathcal{X}_{\mathrm{alt}})$ from -0.2 to 0.6 and $\tau_\text{BDD}(\mathcal{X}_{\mathrm{val}},\mathcal{X}_{\mathrm{alt}})$ from -0.6 to 1.0, without requiring any additional labeling or computationally expensive steps. Meanwhile, it only increases $\tau_\text{KITTI}$ from -0.2 to 0.0 when subsampling from CODA in contrast to $\tau_\text{BDD}$ from -0.6 to 0.6 due to the previously mentioned limited similarity between KITTI and CODA, showcasing the limitations of subsampling. Nevertheless, it proves crucial to match the evaluation set with the deployment domain for a consistent and reliable evaluation of AL methods. The variability in $\tau$ cautions against relying on rankings from a single evaluation set with limited $\mathcal{D_{\text{pool}}}$ coverage. A 1\% difference in mAP, as seen in state-of-the-art comparisons, may not reliably reflect the efficacy of the proposed AL methods.

\begin{figure*}[ht]
  \centering
  \includegraphics[width=\textwidth]{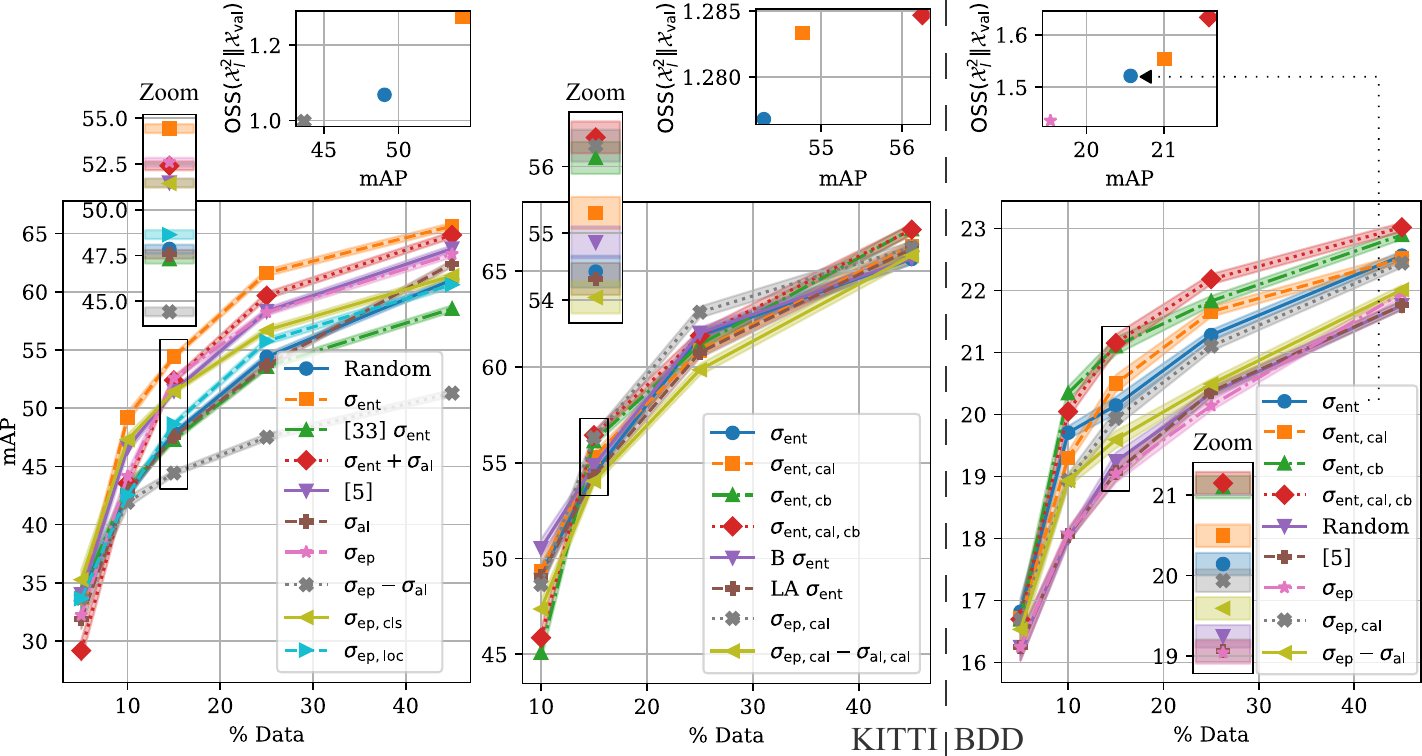}
    \caption{Bottom: Impact of various uncertainties (entropy $\sigma_{\text{ent}}$, epistemic $\sigma_{\text{ep}}$, aleatoric $\sigma_{\text{al}}$, localization $\sigma_{\text{loc}}$ and classification $\sigma_{\text{cls}}$), uncertainty calibration ($\sigma_{\text{cal}}$), class balancing ($\sigma_{\text{cb}}$), and the separation hypothesis (\cref{sec:strat}) on the effectiveness of AL across different iterations ($i$) with a zoom-in on $i=2$. Top: $\mathrm{OSS}$ vs.~mAP of AL methods at $i=2$, showing that performance gains through calibration and class balancing are due to a natural alignment of the selected set with $\mathcal{X}_{\mathrm{val}}$ (as measured by $\mathrm{OSS}$). This differs from the separation hypothesis since the validation set contains high aleatoric uncertainty samples.}
    \label{fig:perfc}
\end{figure*} 

\subsection{Case Study: Uncertainty-Based AL} We complete the analysis on AL by investigating the intuitions in \cref{sec:intro,sec:relwork} and resulting AL methods (see \cref{sec:strat}) through a real-world case study. We observe in \cref{fig:perfc} that the 5\% warm-up has a negligible effect, cf. \cite{chandra2021initial}, as AL methods converge despite $\pm 7\%$ mAP initial performances. $\sigma_\text{ent}$ consistently outperforms standard uncertainty-based methods across all iterations (up to +7\% mAP over \cite{choi2021active} and random), a conclusion supported by previous works \cite{riedlinger2022towards,moses2022localization}. This holds regardless of whether $\sigma_\text{ent}$ stems from a model trained with MC+LA ($\sigma_\text{ent}$), LA only, or the baseline (B) without uncertainty estimation ($<0.5$\% mAP deviation), avoiding the need for costly epistemic uncertainty estimation. This also indicates that the classification confidence of the detector can effectively approximate its epistemic uncertainty without localization information.

However, \cref{fig:perfc} (mid and right) shows that this may be caused by the quality of the uncertainty at $\leq 25\%$ data, as calibrating the uncertainty consistently improves its performance as a selection strategy. We find no advantage to the aggregation strategies \cite{roy2018deep} (\cref{sec:strat}) or separation hypothesis ($\sigma_\text{ep}- \sigma_\text{al}$, $\sigma_\text{ent}+\sigma_\text{al}$, and \cite{choi2021active}). Since $\mathcal{X}_{\mathrm{val}}$ may contain samples with inherently high $\sigma_{al}$, removing them does not improve performance. This shows that AL for real-world applications requires domain-specific adaptation, as theoretically sound methods may not always translate and approaches that work on benchmarking datasets do not necessarily generalize. \cref{fig:perfc} (top) reveals that favoring images with high epistemic but low aleatoric uncertainty decreases similarity. Meanwhile, class balancing ($\sigma_\text{ent,cb}$) proves particularly impactful on both datasets (+1-1.5\% mAP over $\sigma_\text{ent}$) with a stronger effect on BDD due to the stronger class imbalance (see class proportions in \cref{tab:simvsap}). 

We observe that $\sigma_\text{ent,cal,cb}$ performs best (+1.5-2\% mAP over $\sigma_\text{ent}$). The latter can be retraced to both class balancing and calibration inherently increasing the informativeness of the selected samples, as they naturally align the selected set more closely with $\mathcal{X}_{\mathrm{val}}$. This is demonstrated by the higher $\mathrm{OSS}(\mathcal{X}_l^2 \| \mathcal{X}_{\mathrm{val}})$ in the similarity plots of \cref{fig:perfc} (top), further proving that similarity is indeed linked to informativeness. Overall, the performance of AL at 45\% approaches that of 100\%, with \(70.05 \pm 0.23\) on KITTI and \(23.67 \pm 0.13\) on BDD. This highlights the potential of AL for real-world applications, especially with its strategies unified under our similarity-based approach.

\textbf{Method Elimination Simulation:} To demonstrate the practical benefits of $\mathrm{OSS}$, we simulate an $\mathrm{OSS}$-based elimination of AL methods. While \cref{fig:perfc} presents mAP results, obtaining these results requires training detectors first, then evaluating them. The challenge for AL method developers is waiting for training and evaluation to complete to determine if a method is effective, as this is a purely empirical process. Each training takes the times shown in \cref{tab:elimsim}, plus an additional 20 minutes for each evaluation on BDD with a V100 GPU. We evaluate every 20 epochs during training to monitor convergence, adding around three GPU hours per training session. 

\cref{tab:elimsim} presents computational savings achievable through $\mathrm{OSS}$-based method elimination ($S_\mathrm{OSS}$) compared to mAP-based elimination ($S_\text{mAP}$). For KITTI, this approach saves up to $S_\mathrm{OSS}=204$ GPU hours per eliminated method (51 GPU hours per training with four training runs for statistical robustness across multiple seeds). For BDD, the savings increase to 3,224 GPU hours per eliminated method via $\mathrm{OSS}$. Assuming a price of 2.48\$ per GPU hour for a V100 \cite{googlecloud2025}, this reduces cost by up to 7995,52\$ per eliminated method via $\mathrm{OSS}$. Meanwhile, elimination based on mAP post-training would save up to 34 GPU hours less on KITTI (9 GPU hours $\times$ 4 training runs at $i=5$) and 932 hours on BDD (233 GPU hours $\times$ 4 training runs at $i=5$), which is up to 2311,36\$ less savings per eliminated method. \cref{fig:perfc} shows that already at the first iteration, low-performing methods typically retain their rank even as iterations proceed on both datasets (e.g., $\sigma_\text{ep}$ and $\sigma_\text{ep}-\sigma_\text{al}$ can be safely eliminated at $i=1$).

\begin{table}[ht]
\centering
\caption{KITTI (top) and BDD (bottom): Training time in GPU hours for the warm-up step ($i=0$) and each of the six AL iterations ($i\in[1,6]$), including savings if stopping is activated at early iterations based on mAP post-training ($S_\text{mAP}$) and $\mathrm{OSS}$ pre-training ($S_\mathrm{OSS}$).}
\label{tab:elimsim}
\resizebox{\columnwidth}{!}{%
\begin{tabular}{lccccccc}
& \rotatebox{45}{$i=0$} & \rotatebox{45}{$i=1$} & \rotatebox{45}{$i=2$} & \rotatebox{45}{$i=3$} & \rotatebox{45}{$i=4$} & \rotatebox{45}{$i=5$} & \rotatebox{45}{$i=6$} \\
\midrule
Training        & 03    & 04   & 04   & 05    & 06   & 09   & 23   \\
\midrule
$S_\text{mAP}$ &    & 47 & 43 & 38 & 32 & 23 &  \\
$S_\mathrm{OSS}$ &    & \textbf{51} & \textbf{47} & \textbf{43} & \textbf{38} & \textbf{32} &  \\
\midrule
\midrule
Training        & 25 & 29 & 44 & 54 & 164 & 233 & 282 \\
\midrule
$S_\text{mAP}$ &    & 777 & 733 & 679 & 515 & 282 &  \\
$S_\mathrm{OSS}$ &    & \textbf{806} & \textbf{777} & \textbf{733} & \textbf{679} & \textbf{515} &  \\
\bottomrule
\end{tabular}%
}
\end{table}

\subsection{Ablation Studies}
\subsubsection{$\mathrm{OSS}$ Component Analysis}
Our first ablation study focuses on the impact of individual terms in \cref{eq:goeq}, with the results summarized in \cref{tab:acewts}.

\textbf{Main Components:}  Our analysis reveals that the correlation between $\mathrm{OSS}$ and mAP is significantly impacted by the class count term $r_j^{(P_m)}$ in \cref{eq:goeq}, as its removal ($-$) inversely affects the correlation. Moreover, utilizing only $n_j^{(P_m)}$ highlights the importance of a suitable metric comparison (JSD). This is particularly evident in KITTI due to less significant count differences between the comparison sets. Removing smoothing ($T_1$) lowers correlation in three use-cases, confirming its usefulness. Overall, using all terms of \cref{eq:goeq} results in the strongest correlation between $\mathrm{OSS}$ and mAP.
\begin{table}[ht]
    \centering
    \caption{Impact of removing ($-$) each term in \cref{eq:goeq} and using class count as is on the correlation between $\mathrm{OSS}$ and mAP via $r$.}
    \label{tab:acewts}
        \resizebox{\columnwidth}{!}{%
\begin{tabular}{l|c|c|c|c|c}
\toprule
 & All & $r_j^{(P_m)}$ & $n_j^{(P_m)}$ &JSD & $T_1$  \\ \midrule
 & $\checkmark$  & $-$ & $\checkmark$ &$-$& $-$  \\ \midrule
$r_{\mathrm{BDD}}(\mathcal{X}_l^1 \| \mathcal{X}_{\text{alt}})$ & $\textbf{0.81}^{***}$ & $0.45$ & $0.61$ & $0.80^{***}$ & $0.78^{**}$ \\[4pt]
$r_{\mathrm{BDD}}(\mathcal{X}_l^3 \| \mathcal{X}_{\text{alt}})$ & $\textbf{0.89}^{***}$ & $0.59$  & $0.70^*$ & $0.82^{***}$ & $\textbf{0.89}^{***}$ \\[4pt]
$r_{\mathrm{KITTI}}(\mathcal{X}_l^1 \| \mathcal{X}_{\text{val}})$ & $\textbf{0.74}^{**}$ & $-0.21$ & $-0.59$ & $0.63^*$ & $0.69^*$ \\[4pt]
$r_{\mathrm{KITTI}}(\mathcal{X}_l^3 \| \mathcal{X}_{\text{val}})$ & $\textbf{0.83}^{***}$ & $0.35$ & $-0.35$ & $0.65^*$ & $0.76^{**}$ \\[4pt]
\bottomrule
\end{tabular}}
\end{table}  

\textbf{Feature Selection:} The second ablation study expands on the impact of different features on the correlation with performance. \cref{tab:mtrcs} shows that removing any of the three features reduces correlation. Including other image-based features in the JSD such as hue, saturation (SAT), value (VAL), average and standard deviation of pixel intensity, average pairwise structural similarity index measure, and average pairwise Hamming distance of perceptual hashes also prove to be less effective. Switching to model-based features with the average cosine similarity (CS) reverses the correlation and increases computation time. We use for the latter a pre-trained VGG16 model on ImageNet with an output of shape $512\times7\times7$ and an input of shape $224\times224$ \cite{
simonyan2015deep, vgg16}. While $\mathcal{X}_l^1$ of KITTI consists of 1300 crops, there are 128763 crops at $i=1$ for BDD. Calculating CS scales quadratically in contrast to our image-based features distributions. Using CS without our count ratio from \cref{eq:goeq} further reduced the correlation to $r=-0.35$\hyperref[sigfootnote]{\footnotemark[\value{footnote}]} compared to the results in \cref{tab:mtrcs}, which also underscores the efficacy of \cref{eq:goeq}.

\begin{table}[ht]
    \centering
    \caption{Impact of features in \cref{eq:goeq} on $r_{\mathrm{KITTI}}(\mathcal{X}_l^1 \| \mathcal{X}_{val})$ via removing ($-$), adding ($+$) more features to the JSD, and using model-based features with CS.}
    \label{tab:mtrcs}
        \resizebox{\columnwidth}{!}{%
\begin{tabular}{l|c|c|c|c|c|c|c|c}
\toprule
 & JSD & AR & CH & DCT & HUE & SAT & VAL & CS\\ \midrule
 & $\checkmark$ & $-$ & $-$ & $-$& + &+ & + & $\checkmark$  \\ \midrule
$r$ & $\textbf{0.74}^{**}$ & $0.68^*$ & $0.72^*$ & $0.66^*$ & $\textbf{0.74}^{**}$ & $0.71^*$ & $0.69^*$ & $-0.06$ \\

\bottomrule
\end{tabular}}
\end{table} 

\textbf{Balancing Terms:} The third ablation study is on the balancing terms in \cref{eq:reg,eq:w}. Setting $C_w$ around 1.0 activates class weighting through $w_{j}$ exclusively for the strongly class-imbalanced BDD. This strengthens the correlation between similarity and $\tau$ and results in higher $\tau$ post-subsampling (see \cref{tab:balas}). In contrast, removing $w_{j}$, i.e., $w_{j}=1$ for BDD or applying it on KITTI decreases correlation. This highlights the importance of only incorporating the class balancing terms $w_j$ when necessary. Moreover, removing $\beta^{(P_m)}$ causes the similarity for $\mathcal{X}_{\mathrm{val}}$ to dominate over the subsets due to the 1/10 count factor. This leads to its selection as $\mathrm{OSS}_z$ (see \cref{sec:simalrep,fig:evalall}) and therefore a significant drop in the correlation and post-subsampling $\tau$.

\begin{table}[ht]
    \centering
    \caption{Impact of balancing terms in \cref{eq:goeq} on evaluation reliability, measured by $\tau$ post-subsampling.}
    \label{tab:balas}
        \resizebox{0.9\linewidth}{!}{%
\begin{tabular}{l|c|c|c|c|c|c}
\toprule
 & All & $\beta^{(P_m)}$ & $w_j$ & \multicolumn{3}{c}{$C_w$} \\ 
 \midrule
 & $\checkmark$ & $-$ & $-$& 0.7 & 1.0 & 1.4 \\ 
\midrule
$\tau_{\mathrm{BDD}}(\mathcal{X}_{\mathrm{val,subset}} , \mathcal{X}_{\mathrm{alt}})$ & \textbf{0.80} &$-0.60$& $-0.60$& \textbf{0.80} &\textbf{0.80}& $-0.66$\\ 
$\tau_{\mathrm{KITTI}}(\mathcal{X}_{\mathrm{val,subset}} , \mathcal{X}_{\mathrm{alt}})$ & \textbf{0.60} &$-0.20$& -& $-0.22$ &\textbf{0.60}&\textbf{0.60}\\ 
\bottomrule
\end{tabular}}
\end{table} 

\subsubsection{Cross-Detector Validation} The fourth ablation study evaluates the impact of detector choice on $\mathrm{OSS}$. We replace EfficientDet with YOLOv3 \cite{farhadi2018yolov3,zhang2019yolov3tf2}, fine-tuned on KITTI (50 epochs, $1024\times1024$ resolution, batch size 8, learning rate 0.0001, k-means optimized anchors). The strong linear correlation between $\mathrm{OSS}$ and mAP holds across detectors, with average $r_{\mathrm{KITTI}}(\mathcal{X}_l^i \| \mathcal{X}_{\text{val}})=0.92^{***}$ for YOLOv3 ($r=0.97^{***}$ for EfficientDet; see \cref{fig:simap}), confirming $\mathrm{OSS}$ is detector-agnostic despite YOLOv3's lower mAP (1–-9\%).


\section{Conclusion}
We introduce a similarity-based approach that unifies training and evaluation strategies in AL by defining and quantifying informativeness and representativeness. Our approach addresses three key challenges in real-world AL: reducing computational costs, improving evaluation reliability and generalizability despite domain shift, and understanding informativeness in AL from a practical perspective with a case study using uncertainty-based AL methods. We find that uncertainty contributes to informativeness only when it is calibrated and considers the class distribution, thereby naturally increasing similarity between the selected training set and the target domain. Comprehensive experiments on KITTI, BDD100K, and CODA demonstrate the effectiveness of our novel $\mathrm{OSS}$ metric across diverse autonomous driving scenarios

Ultimately, our approach improves the practical usability of AL at minimal computational cost, supporting broader adoption of AL in real-world applications.

\section{Discussion and Future Work}
Most AL works focus on minimizing labeling efforts. Although reducing labeling costs is the primary objective, the required training runs during AL method development are also computationally expensive, an objective often dismissed in previous works. $\mathrm{OSS}$ mitigates this by leveraging already labeled sets to eliminate ineffective AL methods at each iteration. Nevertheless, we also contribute to minimizing labeling efforts in this work by improving uncertainty-based AL methods through uncertainty calibration and class balancing. Future work could explore diversity-based AL and pseudo-labeling in semi-supervised learning as case studies for $\mathrm{OSS}$. We use CODA as an alternative evaluation set to address dataset-specific biases and ensure robust evaluation. Future work could explore other sets with varying levels of domain shift to further understand their effects on evaluation reliability. Additionally, while $\mathrm{OSS}$ finds a reliable evaluation set through subsampling, future work could investigate generative upsampling. The correlation between mAP and $\mathrm{OSS}$ relies on sufficient diversity in class characteristics, struggling with less descriptive rare classes. Future work could use class-balancing augmentations and employ foundation models for more refined feature extraction. While $\mathrm{OSS}$ is not used in this work to select samples, future work could investigate using similarity instead of uncertainty for optimal sample selection. 

\bibliographystyle{IEEEtran}
\bibliography{main}
\end{document}